\documentclass[runningheads]{llncs}
\usepackage[T1]{fontenc}
\usepackage{graphicx}
\usepackage{amsmath}
\usepackage{amssymb}
\usepackage{booktabs}
\usepackage{url}
\usepackage{hyperref}
\usepackage{orcidlink}
\usepackage{microtype}
\hypersetup{colorlinks,linkcolor=blue!60!black,citecolor=blue!60!black,urlcolor=blue!60!black,
  pdftitle={Who Speaks Matters: Authority-Aware Multi-View RAG over Italian Parliamentary Proceedings},
  pdfauthor={Mirko Tritella, Riccardo Pozzi, Matteo Palmonari}}

\begin{document}
\title{Who Speaks Matters: Authority-Aware Multi-View RAG over Italian Parliamentary Proceedings\thanks{This preprint has not undergone peer review (when applicable) or any post-submission improvements or corrections. The Version of Record of this contribution is published in \emph{The Semantic Web -- ISWC 2026}, and is available online at \mbox{\url{https://doi.org/10.1007/978-3-032-42029-9_24}}.}}
\titlerunning{Authority-Aware Multi-View Parliamentary RAG}
\author{Mirko Tritella\inst{1}\orcidlink{0009-0000-8611-8189} \and
Riccardo Pozzi\inst{1}\orcidlink{0000-0002-4954-3837} \and
Matteo Palmonari\inst{1}\orcidlink{0000-0002-1801-5118}}
\authorrunning{M. Tritella et al.}
\institute{University of Milano-Bicocca, Milan, Italy\\
\email{riccardo.pozzi@unimib.it}}
\maketitle
\begin{abstract}
Parliamentary proceedings are a primary record of democratic deliberation, yet their volume and fragmentation make multi-perspective access difficult for citizens, journalists, and researchers. Applying Retrieval-Augmented Generation (RAG) to parliamentary transcripts introduces three specific risks: dominance of the most frequent speakers, inability to weight speakers according to topical expertise, and citation misattribution in politically sensitive text. We present ParliamentRAG, a RAG system for the Italian Chamber of Deputies that addresses these risks jointly. Its core contribution is a topic-dependent authority model that estimates each speaker's authority as a function of the current query, combining interpretable components such as profession, education, and previous interventions. Given a user query, the system retrieves relevant speech chunks, identifies topic-relevant experts across parliamentary groups, and generates a summary synthesizing their perspectives, accompanied by supporting quotations. ParliamentRAG is evaluated against Google NotebookLM on 15 policy topics via a two-level protocol combining automated metrics and blind A/B human evaluation by six domain experts. The system achieves higher coverage across political groups (0.97 vs. 0.95), perfect quotation faithfulness (1.00 vs. 0.95), and stronger expert preferences on source-related dimensions, while NotebookLM remains stronger on prose-oriented dimensions.
\end{abstract}

\keywords{Retrieval-Augmented Generation \and Parliamentary NLP \and Expert Finding \and Knowledge Graphs \and Quotation Faithfulness \and Multi Perspective Summarization.}
\section{Introduction}
\label{sec:intro}
Parliamentary debates constitute a dense documentary record of democratic deliberation: every floor speech, legislative act, and committee intervention expresses the position of an elected representative on matters of public concern. At the time of our experiments (February 2026), the ongoing XIX Legislature of the Italian Chamber of Deputies (October 2022--present) has already produced 608 plenary sessions, 6,010 debates, and 40,416 individual speeches

Manual navigation of such a corpus is excessively costly, and keyword-based institutional search portals return isolated documents rather than synthesized multi-perspective answers. Furthermore, a representative's position on a given topic typically emerges across multiple sessions and distinct debates rather than within a single intervention, compounding the difficulty of reconstruction.

Journalists, researchers, civil society organizations, and citizens increasingly require tools capable of exploring parliamentary activity in a structured and interpretable way, enabling them to reconstruct political positions, monitor legislative behavior, and compare the viewpoints of different parliamentary groups and representatives. Facilitating access to these records is particularly important for citizens, for whom parliamentary debates constitute a primary source of information about how elected representatives discuss public issues, justify political decisions, and position themselves on matters that directly affect society. However, effective access to parliamentary debates requires more than document retrieval alone: systems mediating access to institutional records should provide representative and multi-perspective summaries, preserve attribution and traceability to original speeches, and avoid over-representing isolated interventions or highly active speakers.

Large Language Models (LLMs) are well suited for synthesis, but when applied naively to political text they exhibit three risks in a civic-information settings: (i) they over-represent the most vocal actors, reflecting the distributional asymmetry of the corpus rather than the political landscape of the debate~\cite{10.5555/3618408.3619652}; (ii) they flatten all speakers to content-based similarity, ignoring the rich metadata that distinguishes a topical expert from a tangential commenter; and (iii) they hallucinate or misattribute citations at a rate that, although low, is incompatible with journalistic or institutional use~\cite{Ji_2023,maynez2020faithfulnessfactualityabstractivesummarization}.

In this paper, we present \emph{ParliamentRAG}, a novel RAG application to help both politics professionals (e.g., journalists or analysts) and Italian citizens have a faithful, multi-view, and balanced summary of positions expressed by the ten parliamentary groups of the Italian Chamber of Deputies\footnote{\url{https://www.camera.it/leg19/46}} during parliament debates on a specific topic. Knowledge Graphs (KG) play a crucial role in the development of the application at different levels: all data are stored as a KG because of the rich relational information (e.g., Members of Parliament (MPs) membership of parliamentary groups or MPs signatories for parliamentary acts); graph-based retrieval is a component of the RAG system; the key authority-based ranking method proposed in our paper depends explicitly on graph information.

Our main objective is to generate synthetic summaries of parliamentary debates while respecting three fundamental principles: (i) \textbf{multi-view representation}, ensuring that all parliamentary groups are represented rather than over-relying on frequent speakers; (ii) \textbf{authority awareness}, modeling topical authority dynamically so that the most relevant voices emerge per query; and (iii) \textbf{quotation traceability}, requiring that every quotation corresponds to a verbatim passage from an official parliamentary transcript. These principles aim to support transparent and balanced access to parliamentary deliberation.

\begin{figure}[htbp]
    \centering
    \includegraphics[width=0.7\textwidth]{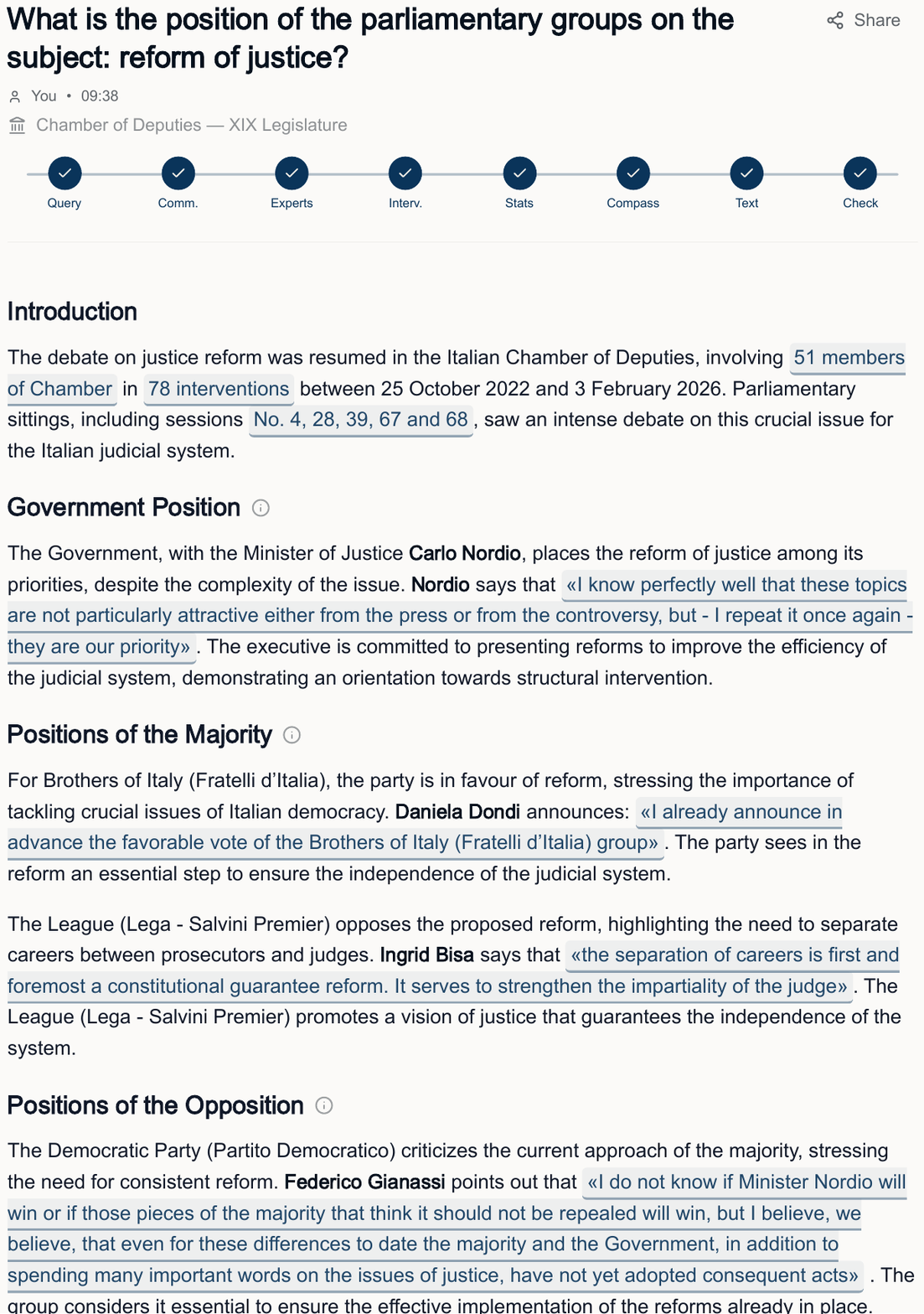}
    \caption{An example of ParliamentRAG's authority-aware, multi-view summary of the positions of the Italian parliamentary groups on the subject \textit{reform of justice}. Exact citations are delimited by << and >>. The image presents the first part of a longer answer in Italian, automatically translated into English for this publication.}
    \label{fig:example}
\end{figure}

Figure~\ref{fig:example} illustrates an authority-aware, multi-view summary of Italian parliamentary positions on justice reform, including traceable verbatim quotations.

\vspace{-5pt}

We operationalize these principles into measurable criteria and evaluate ParliamentRAG through a two-level protocol. First, automated metrics assess structural properties of the generated summaries, including coverage and balance of parliamentary groups, citation fidelity, and alignment between cited speakers and the topic-dependent authority model, providing a scalable and reproducible validation of design compliance.

Second, we conduct a blinded expert survey with six domain experts (parliamentary journalists, collaborators, and policy analysts), who evaluate outputs along qualitative dimensions such as clarity, completeness, perceived balance, and overall usefulness, as well as providing pairwise preferences.

We compare ParliamentRAG against NotebookLM, a strong commercial baseline for document-grounded generation. ParliamentRAG achieves perfect citation faithfulness (1.00 vs.~0.95), near-perfect group coverage (0.97 vs.~0.95), and is consistently preferred on source-related dimensions, particularly Source Coverage (5.7:1 preference ratio). While NotebookLM is preferred on prose-quality aspects, overall satisfaction is comparable, indicating that ParliamentRAG matches a strong baseline while excelling on the dimensions aligned with its architectural design.

A live deployment is available at \url{https://www.parliamentrag.it/} and the source code at \url{https://github.com/Emeierkeio/thesis-ParliamentRAG} under Apache-2.0 license. 

\section{Related Work}
\label{sec:related}

We identify five research threads closely related to the contributions of this work: (i) Natural Language Processing (NLP) and (ii) Knowledge Graphs (KGs) applied to the parliamentary domain, (iii) expert and authority modeling in information retrieval, (iv) fairness- and diversity-aware ranking, and (v) faithful attribution and grounded generation.

To the best of our knowledge, no prior work integrates structured parliamentary graphs with query-dependent authority modeling and multi-view generation within a unified RAG architecture.

\textbf{Parliamentary NLP.} Large parliamentary corpora, including ParlaMint~\cite{erjavec2025parlamint2} and the Italian IPSA corpus~\cite{frasnelli-palmero-aprosio-2024-theres}, have supported tasks such as stance detection, topic modelling, and speaker profiling. Ideological-scaling and embedding-based approaches estimate corpus-level political positions~\cite{slapin2008scaling,Rheault_Cochrane_2020}, but do not support per-query retrieval or multi-view synthesis. More recent LLM- and RAG-based systems have improved access to parliamentary material through faceted search for the Maltese Parliament~\cite{Azzopardi2025} and hybrid vector--KG retrieval for German Bundestag debates~\cite{Mosbach2025}. However, these systems do not jointly address the principles central to our work: multi-view representation, authority-aware retrieval, and quotation-level traceability (Section~\ref{sec:intro}).

\textbf{Knowledge Graphs in the Parliamentary Domain.}
Parliamentary Knowledge Graphs and Linked Open Data initiatives have been developed to improve transparency, interoperability, and large-scale analysis of legislative proceedings. Examples include LinkedEP for European Parliament debates~\cite{vanaggelen2017linkedep}, ParliamentSampo for Finnish parliamentary records~\cite{hyvonen2025publishing}, LinkedSaeima for the Latvian parliament~\cite{bojars2019linkedsaeima}, and the DemocraSci Knowledge Graph for the Swiss Parliament~\cite{salamanca_kg_2024,brandenberger2024democrasci}. In Italy, the Chamber of Deputies exposes legislative data through the OCD ontology and the official \url{dati.camera.it} portal~\cite{daticamera2026}. However, these resources are mainly designed for structured querying, exploration, and data integration, rather than for query-dependent reasoning, multi-view generation, or authority-aware interpretation. Our work addresses this gap by integrating parliamentary KG data into a retrieval-augmented generation architecture for multi-view and authority-aware inference (Section~\ref{sec:kg}).

\textbf{Expert Finding and Authority.} Expert finding formalises the identification of authoritative individuals from text and metadata~\cite{balogexpertise}, distinguishing profile-based from document-based approaches. Link-based ranking~\cite{Page1998PageRank,10.1145/324133.324140} yields query-independent authority. Source-reliability estimation in RAG~\cite{hwang2024retrieval} assumes a single factually correct answer, an assumption that does not hold in parliamentary debate, where multiple legitimate viewpoints coexist. No prior system computes \emph{interpretable, query-dependent} authority from observable parliamentary features while modelling temporal decay and coalition-crossing invalidation.

\textbf{Multi-View Retrieval and Fairness.} Fairness-aware ranking~\cite{Zehlike_2017} and diversification methods such as MMR~\cite{10.1145/290941.291025} optimise for content or demographic diversity but are not integrated into RAG pipelines for political applications. GraphRAG~\cite{edge2025localglobalgraphrag} exploits document structure but does not enforce coverage of predefined groups.

\textbf{Quotation Faithfulness.} Attribution frameworks~\cite{rashkin-etal-2023-measuring}, post-hoc systems such as RARR~\cite{gao-etal-2023-rarr}, and verification methods~\cite{liu2023evaluatingverifiabilitygenerativesearch} improve quotation quality, but the generator remains free to paraphrase the source. Our ParliamentaryRAG system, instead, provides a \emph{by-construction} guarantee that every quotation is a verbatim substring of a specific source document. 

\section{Parliamentary Knowledge Graph}
\label{sec:kg}

ParliamentRAG operates over a knowledge graph (KG) that integrates parliamentary proceedings, metadata, and legislative activity into a unified, Italian-language representation. The underlying data is sourced from the open data infrastructure of the Italian Chamber of Deputies~\cite{daticamera2026}, which publishes semantically structured records in RDF format according to the \emph{Ontology of the Chamber of Deputies}.\footnote{\url{https://dati.camera.it/ocd/classi.rdf}} The dataset includes plenary and committee transcripts, legislative acts with signatories, deputy biographical information, and committee memberships with temporal validity.

\begin{figure}[t]
  \centering
  \includegraphics[width=0.7\linewidth]{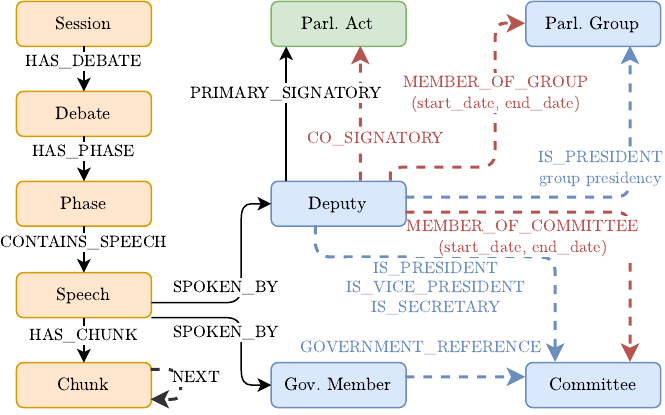}
  \caption{Knowledge graph schema stored in a single Neo4j instance.}
  \label{fig:knowledge-graph}
\end{figure}

The RDF data is transformed into a property graph stored in a single Neo4j\footnote{\url{https://neo4j.com/}} instance, which serves as the sole data layer for the entire system. This architectural choice is motivated by three factors: (i)~the need to represent relationships with properties (e.g., temporal validity such as \texttt{start\_date} and \texttt{end\_date} on \texttt{MEMBER\_OF\_GROUP} and \texttt{MEMBER\_OF\_COMMITTEE}; see Figure~\ref{fig:knowledge-graph}), which are naturally handled in a property graph model; (ii)~the integration of vector similarity search within the same system, avoiding the overhead of a separate vector store; and (iii)~the ability to combine graph traversal, keyword search, and dense retrieval within a unified query infrastructure.

During transformation, all entities and relationships from the source ontology are preserved. The graph includes 387 deputies (out of 400 elected members; 13 never intervened in plenary debate and thus have no associated speeches), 64 government members (only those who delivered at least one speech during a Chamber session), 10 parliamentary groups, 80 committees, 608 sessions, 6{,}010 debates, 6{,}515 phases, 40{,}416 speeches, and 27{,}576 legislative acts. The graph is further enriched with information not present in the RDF source: the educational background and profession of MPs, extracted from their short biographical descriptions on the Chamber's website, are embedded and stored as node properties (\texttt{education\_embedding}, \texttt{profession\_embedding}). Semi-structured parliamentary data---debate and session transcripts published in XML format\footnote{\url{https://documenti.camera.it/apps/commonServices/getDocumento.ashx?sezione=assemblea\&tipoDoc=formato\_xml\&tipologia=stenografico\&idNumero=0655\&idLegislatura=19}}---are parsed to extract the hierarchical structure of proceedings and integrated into the graph alongside institutional data. The resulting graph contains 232{,}755 nodes and 488{,}487 relationships across 13 node labels and 15 relationship types.

The overall schema is shown in Figure~\ref{fig:knowledge-graph}. Parliamentary proceedings are organized as a hierarchical structure \textsc{Session} $\rightarrow$ \textsc{Debate} $\rightarrow$ \textsc{Phase} $\rightarrow$ \textsc{Speech} $\rightarrow$ \textsc{Chunk}, mirroring the institutional workflow. Speeches are segmented into 151{,}073 chunks, which serve as the atomic retrieval units. Each chunk stores its textual content, a 1{,}536-dimensional dense embedding (text-embedding-3-small), its position within the speech, and character-level offsets (\texttt{start\_char\_raw}, \texttt{end\_char\_raw}) into the original transcript, allowing retrieved evidence to be traced back to verbatim source text. Sequential chunks are linked by \texttt{NEXT} edges (110{,}657 links), supporting context-window expansion during generation.

Raw transcripts contain both substantive content and procedural elements (e.g., voting announcements or formal statements), which are removed through lightweight preprocessing. To preserve exact traceability, an alignment between cleaned and raw text is maintained, enabling deterministic reconstruction of original passages for quotation grounding.

Beyond the textual hierarchy, the graph captures the political and institutional context of speakers. Deputies are linked to parliamentary groups and committees through time-qualified relationships (505 group memberships, 1{,}515 committee memberships), allowing affiliation to be resolved at any point in time---a capability essential for coalition-crossing detection in the authority scoring (Section~\ref{sec:approach}). They are also connected to legislative acts via \texttt{PRIMARY\_SIGNATORY} (27{,}373) and \texttt{CO\_SIGNATORY} (103{,}959) edges, providing a structural signal of topic engagement complementary to speech content. Institutional roles (\path{IS\_PRESIDENT}, \texttt{IS\_VICE\_PRESIDENT}, \texttt{IS\_SECRETARY}) are recorded for committees and parliamentary groups, and are used as signals in authority estimation.

The graph supports three types of access: dense semantic search over chunk embeddings (via Neo4j's native vector index), keyword search over act titles (via a full-text index), and structural graph traversal. This unified representation lets the system combine text and structure, supporting retrieval (e.g., traversing from a parliamentary act to its signatories and their speeches), authority modelling (e.g., aggregating a deputy's legislative activity across signed acts), and quotation grounding (i.e., linking each generated quote back to the exact source passage via stored character offsets).

\section{ParliamentRAG}
\label{sec:approach}

ParliamentRAG is a web application built on top of Next.js\footnote{\url{https://nextjs.org/}} and FastAPI\footnote{\url{https://fastapi.tiangolo.com/}} that provides an interactive interface for exploring Italian parliamentary proceedings through retrieval-augmented generation. The system is publicly accessible at \url{https://www.parliamentrag.it}.

The interface supports free-text queries over parliamentary debates and presents structured, multi-perspective answers grounded in official transcripts. In addition to the main search functionality, it offers auxiliary tools for targeted exploration, including filtered search over legislative acts and speeches, authority analysis of deputies on specific topics, and visualization of parliamentary positions. 

We now describe the underlying pipeline that transforms a user query into a multi-view, quotation-grounded response, jointly modeling \emph{what is said} and \emph{who is saying it}. The system follows a retrieval--ranking--generation pipeline in which speaker authority, coverage of political groups, and quotation faithfulness are enforced as first-class constraints.

\begin{figure}[t]
  \centering
  \includegraphics[width=.6\linewidth]{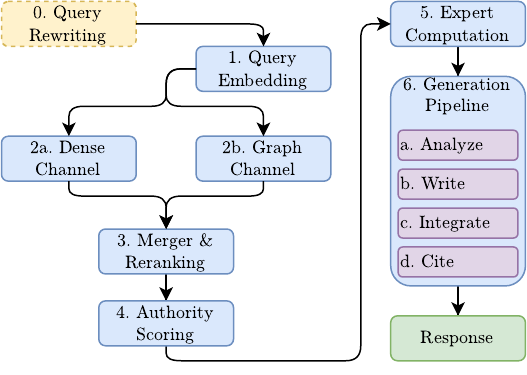}
  \caption{Processing Pipeline. Steps 2a and 2b are executed in parallel.
  }
  \label{fig:processing-pipeline}
\end{figure}

Given a query $q$, the system (i) retrieves candidate evidence from parliamentary proceedings via a dual-channel strategy, (ii) reranks evidence using a composite score that incorporates query-dependent speaker authority and group coverage, (iii) selects representative experts for each parliamentary group, and (iv) generates a structured multi-view answer with verbatim quotations extracted from source documents.
A schema of the pipeline is depicted in Figure~\ref{fig:processing-pipeline}.

\subsection{Dual-channel Retrieval}
The retrieval stage must identify speech chunks that are (i) semantically relevant with respect to the user query, (ii) representative of all parliamentary groups, and (iii) reflective of speakers whose expertise is evidenced by legislative activity rather than solely by speech content.

To this end, we adopt a dual-channel approach. A \emph{dense channel} performs vector similarity search over speech chunks, capturing thematically relevant content independently of the speaker. Later, graph traversal is used to resolve the associated speech, speaker, and parliamentary group, also evaluating if the speaker's current group differs from that at the time of the speech. The primary limitation of the dense channel is that it cannot distinguish between a deputy who has spoken tangentially about a topic and
one who has actively sponsored legislation in the same domain.
The \emph{graph channel} retrieval is based on parliamentary acts. First, acts relevant to the query are selected with hybrid lexical and dense, semantic similarity, then the primary signatories and co-signatories are obtained with graph traversal, as well as their speeches and chunks.

Additionally, when the query can be mapped to a parliamentary committee via a curated keyword-based mapping\footnote{\url{https://github.com/Emeierkeio/thesis-ParliamentRAG/blob/main/backend/config/commissioni_topics.yaml}}, the retrieval engine additionally privileges chunks from speakers who are members of the relevant committee (e.g., justice or foreign affairs; this information is found in the KG), since committee membership is a strong institutional proxy for domain expertise.

The graph channel explicitly considers expert-authored legislative activity. A deputy who has co-signed a relevant act is, by this criterion, likely to have
engaged substantively with the topic. This signal is invisible to the dense channel, which
operates exclusively on chunk-level semantic similarity.
The combination of the two channels yields a more diverse result set, spanning a broader range of political groups and evidence types, than either channel alone.

\subsection{Authority-Aware Reranking and Expert Computation}

Retrieved evidence is merged deduplicating items by chunk identifier and retaining the higher score. Each evidence item $e$ is ranked using a weighted score:
\begin{equation}
\operatorname{score}(e) = w_r r(e) + w_d d(e) + w_v v(e) + w_a \mathit{authority}(s_e, q) + w_\sigma \sigma(e),
\end{equation}
where $s_e$ denotes the speaker of $e$.
The weights are set as follows: relevance $w_r=0.35$, diversity $w_d=0.15$, coverage $w_v=0.20$, authority $w_a=0.05$, and salience $w_\sigma=0.25$. Relevance is prioritized as the primary retrieval objective, while salience promotes meaningful political content over procedural language. Coverage and diversity ensure balanced representation across parliamentary groups and speakers. Authority models speaker expertise with respect to the query. Its weight is the smallest to inform ranking without overriding topical relevance, since a higher authority weight would risk marginalizing speakers from small parliamentary groups who may lack senior
institutional roles but have made substantive contributions to the debate.

The authority is computed after retrieval. This design ensures that authority informs
but does not determine evidence selection, preserving political balance. Authority is also
used for expert selection (one top-authority speaker per group), which influences the
generation pipeline's but not the evidence pool itself.

\textbf{Authority Score.}
We define a query-dependent authority score for each speaker by aggregating multiple signals capturing biographical information, institutional roles, and parliamentary activity. The score combines cosine similarity between query embeddings and the embeddings of speaker attributes (profession, education, committees, and roles) with activity-based signals derived from legislative acts and speech interventions, both subject to temporal decay.

Formally, the raw authority score is computed as a weighted sum of heterogeneous components:
\begin{equation}
\mathit{authority}(s,q) = \sum_i w_i \, c_i(s,q),
\end{equation}
where components include semantic similarity terms, time-decayed counts of parliamentary activity, and role-based priors.

The assigned weights are $w_{\text{profession}}=0.15$, $w_{\text{education}}=0.10$, $w_{\text{committee}}=0.25$, $w_{\text{legislative acts}}=0.20$, $w_{\text{speech interventions}}=0.25$, and $w_{\text{institutional role}}=0.05$. These weights are currently set empirically based on expert judgment and domain knowledge, and are not learned from data. This design choice allows explicit control over the contribution of each component in the authority model, making the scoring function interpretable and adaptable. In particular, the formulation enables future extensions towards personalization or context-specific reweighting strategies.

Authority signals are time-decayed so that recent parliamentary activity contributes more than older evidence, applying the same decay rate to both legislative acts and speech interventions to prioritize current relevance and avoid overemphasizing past prominence.

Coalition affiliation is time-dependent and resolved at query time, ensuring that only activities within a speaker's current political group are considered, preventing historical shifts between coalitions from biasing present-day authority estimates.

\textbf{Expert Computation.}
At this point, for each parliamentary group $g \in G$, the system selects the speaker with the highest query-dependent authority score among those from $g$ present in the evidence pool. This speaker serves as the group's expert representative: their interventions and cited passages become the primary voice through which that group's position is conveyed in the summary.

\subsection{Multi-View Generation Pipeline}
As visible in Figure~\ref{fig:processing-pipeline}, 
the generation process follows a four-stage pipeline---\textit{Analyze, Generate, Integrate, Cite}---that enforces coverage, grounding, and coherence by design. In the first stage (\textit{Analyze}), the input query is split into a set of simple, atomic claims, each linked to the evidence required for every parliamentary group. This ensures that all relevant aspects of the query are explicitly addressed in the following steps.

In the second stage (\textit{Generate}), the system produces one section per parliamentary group from a structured \textit{position brief} summarizing the top evidence chunks for that group, ordered by authority score so that the most authoritative speakers are cited first. All groups are treated uniformly, and if no supporting evidence is available, this is stated explicitly instead of generating unsupported content. The third stage (\textit{Integrate}) combines these sections into a single coherent narrative, while preserving alignment with the original content and ensuring that all quotation placeholders are retained.

In the final stage (\textit{Cite}), quotations are resolved deterministically into verbatim quotations from the parliamentary record. The language model never generates quotation text: it only inserts placeholders with character offsets that are later replaced with the text from the original speeches. As a result, every quote is directly grounded in source text, preventing hallucinations by construction and ensuring that all claims are supported by verifiable evidence.

\section{Experimental Evaluation}
\label{sec:eval}

The experimental evaluation focuses on the downstream quality of generated responses and assesses whether ParliamentRAG effectively supports the exploration of parliamentary debates and policy positions while satisfying the core design principles introduced in Section~\ref{sec:intro} (balanced multi-view coverage across parliamentary groups, authority-aware evidence selection, grounded and traceable quotations). More specifically, we address the following research question: \emph{(i) what is the quality of the analyses produced by ParliamentRAG, a RAG architecture explicitly designed around the principles introduced above, in light of these design principles; and (ii) how does this quality compare with that of analyses produced by a top-tier general-purpose document-grounded RAG system explicitly instructed to follow the same principles?}

For our comparison, we selected a top-tier general-purpose source-grounded system, namely Google NotebookLM\footnote{\url{https://notebooklm.google.com/}}, used in a controlled settings (see Section~\ref{sec:evaluationprotocol}) for different reasons: we are not aware of other competing systems evaluated and published in peer-reviewed publications; NotebookLM represents a realistic and accessible alternative that end users could readily employ to analyze parliamentary material without developing a custom pipeline, after they autonomously collect the data; it is a highly engineered system backed by the top-tier Gemini 3.0 LLM~\footnote{\url{https://x.com/NotebookLM/status/2002115447425282449}} used by hundred millions of users. 

To answer our research questions, we consider a response-level automatic analysis with a qualitative assessment of the perceived quality by domain experts with a blind A/B test on a benchmark of questions. Automatic evaluation metrics focus on the satisfaction of our design principles, such as coverage balance, quotation faithfulness, and adherence to the intended analytical principles, while experts' judgments focus on different aspects of perceived quality and overall satisfaction.
We describe the benchmark construction and evaluation methodology in the following sections and discuss experimental findings afterward.

\subsection{Benchmark and Protocol}
\label{sec:evaluationprotocol}
\textbf{Selection of benchmark topic-based questions.} A first step consists in identifying topics of interest to guide the actual test. We collected 51 policy topics through a survey involving 17 Italian citizens interested in politics, and reduced them to 15 by selecting topics spanning different levels of perceived polarization and public discussion intensity, as assessed directly in the survey.
Topics span eight thematic macro-areas: Justice \& Civil Rights, Institutional Reforms, Immigration, Labour \& Welfare, Environment \& Energy, Technology, Defence \& Foreign Policy, Economy \& Finance, with varying debate intensity and polarization (Likert 1--5). For each topic we create a query with the following template: ``Qual è la posizione dei gruppi parlamentari su \{topic\}?'' (Italian for ``What is the position of the parliamentary groups on \{topic\}?'').

\textbf{NotebookLM as a baseline.} ParliamentRAG operates the retrieval step on all 151k chunks extracted from the parliament proceedings. Feeding all proceedings data to NotebookLM, a general-purpose system, is unfeasible and unfair because the system is not engineered to work with this kind and size of data. We therefore create a notebook for each topic by engineering its context to simulate the case where an expert selects proceedings deemed of interest for a topic; observe that this selection step is challenging, as a topic can be addressed across different proceedings, while ParliamentRAG retrieves data from the whole proceeding data. We select $\approx 150$ relevant $+\ 150$ distractor chunks per topic-specific notebook. Chunks are enriched with metadata to provide additional context (speaker, parliamentary group, date, session, debate). 
The relevant ones are obtained starting from the query with ParliamentRAG retrieval pipeline, including the reranking, which considers the authority score, while the distractors are obtained from similar topics (same macro-area but different topic). Distractors provide a fairer simulation of retrieval-based context and the presence of non-relevant information in uploaded documents. 

\textbf{Comparison of ParliamentRAG and NotebookLM.} A key distinction between the two systems lies in how they are guided towards the design principles: ParliamentRAG incorporates them directly into its architecture through explicit mechanisms, such as authority-aware ranking and multi-view retrieval across parliamentary groups. By contrast, NotebookLM is guided through prompt instructions, without architectural guarantees; all prompts are available in the code base\footnote{\url{https://github.com/Emeierkeio/thesis-ParliamentRAG/blob/main/Prompts.md}}. 
The comparison therefore evaluates whether structurally enforced constraints lead to more reliable and informative outputs than prompt-level control alone. Another important difference concerns the backbone LLMs: ParliamentRAG uses GPT-4o, whereas NotebookLM used Gemini 3 at the time of the experiments. Gemini 3 models are more recent (November 2025\footnote{\url{https://blog.google/products-and-platforms/products/gemini/gemini-3/}}) and outrank GPT-4o, released in May 2024\footnote{\url{https://openai.com/index/hello-gpt-4o/}}, in public rankings\footnote{\url{https://arena.ai/leaderboard/text}}~\cite{chatbotarena2024}. 
In summary, both systems are instructed to produce per-group structured responses via equivalent prompts, while NotebookLM is given an engineered context and backed by a stronger LLM. This setup provides a conservative comparison for ParliamentRAG, although it should be interpreted primarily as an evaluation of response quality under curated retrieval conditions rather than as a fully independent end-to-end retrieval comparison.

\textbf{Automated Evaluation Protocol.} To measure whether the generated answers satisfy the core principles of ParliamentRAG, we first calculate automatic metrics. 
Coverage is evaluated through two complementary measures. \emph{Groups with Quotation (GQ)} quantifies the fraction of the ten parliamentary groups for which at least one speaker has been quoted in the response, while \emph{Completeness} measures the fraction of groups receiving a dedicated narrative section in the generated analysis. To evaluate quotation reliability, we compute \emph{Quotation Faithfulness}, defined as the fraction of quotations that exactly match substrings of the original parliamentary interventions. Finally, to analyze the behavior of the authority-aware retrieval mechanism, we report the mean authority score (\emph{MA}) and the corresponding standard deviation (\emph{ASD}) of the cited speakers under the query-dependent authority model.

\textbf{Human Evaluation Protocol.}
Six domain experts, disjoint from the paper authors, were selected among parliamentary journalists, parliamentary collaborators, and policy analysts, with a mean experience of 7.2 years. They independently rated both systems on nine Likert (1--5) dimensions: Answer Quality, Answer Clarity, Answer Completeness, Quotation Relevance, Balance Perception, Balance Fairness, Source Relevance, Source Authority, Source Coverage. Evaluators also provided an overall A/B preference and a 1--5 satisfaction score. The protocol follows the PARADISE framework~\cite{walker1997} and the A/B preference format of Chatbot Arena~\cite{chatbotarena2024}. The evaluation was blind: system identities were hidden and left/right assignment randomized per topic. Two evaluators completed all 15 topics, while other evaluators partially completed the evaluation, yielding a total of $N = 67$ paired evaluations.

We assess differences between the two systems using non-parametric paired tests (Wilcoxon signed-rank) with Holm--Bonferroni correction for multiple comparisons. In addition, we calculate Cohen's $d$ as an effect size measure to quantify the magnitude of observed differences.

\subsection{Results}
\vspace{-1em}

\label{sec:results}
\begin{table}[htbp]
\centering
\caption{Automated Evaluation: ParliamentRAG vs.\ NotebookLM. Mean values ($\mu$) over the 15 topics and difference ($\Delta$). Mean Authority (MA) for NotebookLM is computed from the query-related authority of the MPs mentioned by NotebookLM.}
\label{tab:automated_metrics}
\small
\setlength{\tabcolsep}{5pt}
\begin{tabular}{lccc}
\toprule
\textbf{Metric} & ParliamentRAG ($\mu$) & NotebookLM ($\mu$) & $\Delta$ \\
\midrule
Groups with Quotation (GQ) & $0.97$  & 0.95  & $+0.02$ \\
Completeness & $0.99$  & 1.00  & $-0.01$ \\
Quotation Faithfulness (QF) & $1.00$  & 0.95  & $+0.05$ \\
Mean Authority (MA) & $0.53$  & $0.52$ & $+0.01$ \\
\bottomrule
\end{tabular}%
\end{table}
\vspace{-0.5em}

Table~\ref{tab:automated_metrics} reports the resulting automated metrics. ParliamentRAG achieves near-perfect quotation-level group coverage ($\mathrm{GQ} = 0.97$ vs.~$0.95$ for NotebookLM), which is architecturally guaranteed at generation-time by stratified per-group generation. However, coverage remains bounded by the retrieval stage: if no retrieved chunks are associated with a given political group, that group cannot be represented during generation, explaining why $\mathrm{GQ} \neq 1.0$.
NotebookLM omits one or more groups 5\% of the times. On Quotation Faithfulness the gap is structurally significant: ParliamentRAG reaches $\mathrm{QF} = 1.00$ by design, whereas NotebookLM reaches $0.95$, meaning approximately 5\% of its quotations do not correspond verbatim to the source---an error rate that, in a journalistic context, is corrosive to credibility. Mean authority of cited speakers is modestly higher for ParliamentRAG ($0.53$ vs.~$0.52$). We remind that Mean Authority (MA) for NotebookLM is computed from the query-related authority of the MPs mentioned by NotebookLM and that the relevant chunks are obtained with the same retrieval strategy as ParliamentaryRAG and reranked considering the authority socre; hence, this could cause the only slightly difference of $0.01$.

\begin{table}[htbp]
\centering
\caption{Human evaluation comparing ParliamentRAG (PRAG) and NotebookLM (NbLM). The left section reports mean Likert ratings (1--5) for each evaluation dimension, while the right section reports the percentage of pairwise preferences in favor of PRAG, ties, and preferences in favor of NbLM.}
\label{tab:human_ratings}
\setlength{\tabcolsep}{5pt}
\begin{tabular}{lcc|ccc}
\toprule
Dimension & PRAG ($\mu$) & NbLM ($\mu$) & PRAG Pref. & Ties & NbLM Pref.  \\
\midrule
Answer Quality & 4.04  & 4.30        &  30\% & 25\% & 45\% \\
Answer Clarity & 4.27  & 4.51        &  22\% & 48\% & 30\% \\
Answer Complet. & 4.12  & 4.09   &  24\% & 55\% & 21\% \\
Quotation Relevance & 4.37  & 4.36   &  24\% & 54\% & 22\% \\
Balance Perception & 4.66  & 4.48    &  21\% & 72\% & 7\%\\
Balance Fairness & 4.67  & 4.66      &  12\% & 78\% & 10\% \\
Source Relevance & 4.07  & 3.84      &  33\% & 48\% & 19\% \\
Source Authority & 4.21  & 4.00      &  30\% & 57\% & 13\% \\
Source Coverage & 4.64  & 4.39       &  25\% & 70\% & 5\%\\
Overall Satisfaction & 4.24  & 4.27  &  31\% & 42\% & 27\% \\
\bottomrule
\end{tabular}
\end{table}

Table~\ref{tab:human_ratings} reports the results of the human evaluation. Overall satisfaction is nearly identical across the two systems ($4.24$ vs.\ $4.27$), suggesting comparable perceived utility despite substantial architectural differences. However, the dimension-level analysis reveals complementary strengths.

NotebookLM achieves higher ratings on prose-oriented dimensions, namely Answer Quality ($4.30$ vs.\ $4.04$) and Answer Clarity ($4.51$ vs.\ $4.27$), and is more frequently preferred on Answer Quality (45\% vs.\ 30\%). This likely reflects the stronger fluency and stylistic coherence of a monolithic generation approach powered by a frontier commercial LLM.

By contrast, ParliamentRAG consistently outperforms NotebookLM on dimensions directly related to the objectives of the proposed architecture. In particular, evaluators report higher ratings for Source Relevance ($4.07$ vs.\ $3.84$), Source Authority ($4.21$ vs.\ $4.00$), and Source Coverage ($4.64$ vs.\ $4.39$). Pairwise preferences reinforce this pattern: ParliamentRAG is preferred more frequently on Source Relevance (33\% vs.\ 19\%), Source Authority (30\% vs.\ 13\%), and especially Source Coverage (25\% vs.\ 5\%). Similarly, ParliamentRAG obtains a clear advantage on Balance Perception ($4.66$ vs.\ $4.48$), although the large proportion of ties (72\%) indicates that both systems are often perceived as balanced.

Although none of the observed differences reaches statistical significance after Holm--Bonferroni correction for multiple comparisons, the results exhibit a coherent qualitative pattern. Small but consistent Cohen's $d$ effect sizes favor ParliamentRAG on source- and balance-related dimensions, including Source Relevance ($d = 0.35$), Source Coverage ($d = 0.35$), and Source Authority ($d = 0.28$), while NotebookLM shows advantages on fluency-oriented dimensions such as Answer Quality ($d = -0.31$) and Answer Clarity ($d = -0.29$).

Finally, after completing the per-topic evaluation, respondents were asked whether they would recommend a similar system to their colleagues, with ``no'' set as the predefined answer. They answered affirmatively in 72\% of cases, suggesting that such a system was generally perceived as useful. 

\vspace{-5pt}

\section{Discussion}
\label{sec:discussion}

The downstream evaluation reveals complementary strengths between the two systems. ParliamentRAG consistently performs better on source-oriented and balance-related dimensions, including source authority, source coverage, source relevance, and perceived political balance, while NotebookLM achieves higher ratings on fluency-oriented aspects such as answer quality and clarity. This asymmetry can be explained by ParliamentRAG prioritizing structural guarantees through explicit multi-view retrieval and authority-aware evidence selection. Better fluency and clarity in NotebookLM can be explained either as the outcome of a monolithic generation pipeline or as the result of using a better backbone model. However, the observed complementarity is not symmetric. Fluency limitations can often be mitigated through stronger language models or lightweight rewriting stages, whereas guarantees such as verbatim quotation faithfulness and systematic per-group coverage require architectural support and cannot be reliably enforced through prompting alone. 
It is also important to recall that NotebookLM was evaluated with a curated, pre-retrieved context, rather than over the full parliamentary corpus as ParliamentRAG was. Thus, the comparison does not test NotebookLM's ability to retrieve evidence from the complete proceedings; in practice, this task would remain substantially more challenging for end users without a dedicated system such as ParliamentRAG.

Among the evaluated properties, quotation faithfulness is the strongest invariant enforced by the architecture. Since quotations are extracted through offset-based retrieval directly from parliamentary transcripts, fabricated or altered quotations are structurally prevented. Other properties, such as balanced group coverage, should instead be interpreted as strong design objectives rather than absolute guarantees, as they still depend on the availability of relevant retrieved evidence.

\textbf{Limitations.} The evaluation is limited by the small benchmark size (15 topics) and moderate statistical power. In addition, the study compares the complete system against a strong external baseline without internal ablation experiments, leaving the contribution of architectural components to future work.

\section{Conclusion}
\label{sec:conclusion}

We have presented ParliamentRAG, an authority-aware, multi-view RAG system for Italian parliamentary proceedings. On a 15-topic benchmark with expert A/B evaluation, the system matches a strong commercial baseline (NotebookLM) on overall satisfaction while excelling precisely on the dimensions it was designed to target (group coverage, quotation faithfulness, source authority). Directions for future work include the personalization of ranking considering users' preferences,
domain-specific embedding models, extension to the Senate and to earlier legislatures, and cross-national generalization to European parliamentary families.

\paragraph*{Supplemental Material Statement:} Source code is released at \url{https://github.com/Emeierkeio/thesis-ParliamentRAG} under Apache-2.0 license.
The knowledge graph can be reconstructed from public data (see Section~\ref{sec:kg}).

\begin{credits}
\subsubsection{\ackname} We thank the six domain experts who contributed to the A/B evaluation and the 17 respondents to the participatory topic-collection survey.

\end{credits}

\section*{Use of Generative AI}

Generative AI tools were used during the preparation of this work to assist with language refinement. In particular, these tools were employed to paraphrase selected passages and improve the clarity, fluency, and overall quality of the English text.

\bibliographystyle{splncs04}
\bibliography{bibliography}

@inproceedings{chatbotarena2024,
author = {Chiang, Wei-Lin and Zheng, Lianmin and Sheng, Ying and Angelopoulos, Anastasios N. and Li, Tianle and Li, Dacheng and Zhu, Banghua and Zhang, Hao and Jordan, Michael I. and Gonzalez, Joseph E. and Stoica, Ion},
title = {Chatbot arena: an open platform for evaluating LLMs by human preference},
year = {2024},
publisher = {JMLR.org},
booktitle = {Proceedings of the 41st International Conference on Machine Learning},
articleno = {331},
numpages = {30},
location = {Vienna, Austria},
series = {ICML'24}
}

@article{10.1145/324133.324140,
    author = {Kleinberg, Jon M.},
    title = {Authoritative sources in a hyperlinked environment},
    year = {1999},
    issue_date = {Sept. 1999},
    publisher = {Association for Computing Machinery},
    address = {New York, NY, USA},
    volume = {46},
    number = {5},
    issn = {0004-5411},
    url = {https://doi.org/10.1145/324133.324140},
    doi = {10.1145/324133.324140},
    journal = {J. ACM},
    month = sep,
    pages = {604–632},
    numpages = {29}
}

@inproceedings{10.5555/3618408.3619652,
    author = {Santurkar, Shibani and Durmus, Esin and Ladhak, Faisal and Lee, Cinoo and Liang, Percy and Hashimoto, Tatsunori},
    title = {Whose opinions do language models reflect?},
    year = {2023},
    publisher = {JMLR.org},
    booktitle = {Proceedings of the 40th International Conference on Machine Learning},
    articleno = {1244},
    numpages = {34},
    location = {Honolulu, Hawaii, USA},
    series = {ICML'23}
}

@article{Rheault_Cochrane_2020,
    title={Word Embeddings for the Analysis of Ideological Placement in Parliamentary Corpora},
    volume={28},
    DOI={10.1017/pan.2019.26},
    number={1},
    journal={Political Analysis},
    author={Rheault, Ludovic and Cochrane, Christopher},
    year={2020},
    pages={112–133}
}

@misc{maynez2020faithfulnessfactualityabstractivesummarization,
      title={On Faithfulness and Factuality in Abstractive Summarization}, 
      author={Joshua Maynez and Shashi Narayan and Bernd Bohnet and Ryan McDonald},
      year={2020},
      eprint={2005.00661},
      archivePrefix={arXiv},
      primaryClass={cs.CL},
      url={https://arxiv.org/abs/2005.00661}, 
}

@inproceedings{frasnelli-palmero-aprosio-2024-theres,
    title = "There{'}s Something New about the {I}talian Parliament: The {IPSA} Corpus",
    author = "Frasnelli, Valentino  and
      Palmero Aprosio, Alessio",
    editor = "Calzolari, Nicoletta  and
      Kan, Min-Yen  and
      Hoste, Veronique  and
      Lenci, Alessandro  and
      Sakti, Sakriani  and
      Xue, Nianwen",
    booktitle = "Proceedings of the 2024 Joint International Conference on Computational Linguistics, Language Resources and Evaluation (LREC-COLING 2024)",
    month = may,
    year = "2024",
    address = "Torino, Italia",
    publisher = "ELRA and ICCL",
    url = "https://aclanthology.org/2024.lrec-main.1394/",
    pages = "16037--16046"
}

@article{slapin2008scaling,
  title={A scaling model for estimating time-series party positions from texts},
  author={Slapin, Jonathan B and Proksch, Sven-Oliver},
  journal={American Journal of Political Science},
  volume={52},
  number={3},
  pages={705--722},
  year={2008},
  publisher={Wiley Online Library}
}

@article{erjavec2025parlamint2,
    author = {Erjavec, Toma{\v{z}} and Kopp, Maty{\'a}{\v{s}} and Ljube{\v{s}}i{\'c}, Nikola and Kuzman, Taja and Rayson, Paul and Osenova, Petya and Ogrodniczuk, Maciej and {\c{C}}{\"o}ltekin, {\c{C}}a{\u{g}}r{\i} and Kor{\v{z}}inek, Danijel and Meden, Katja and Skubic, Jure and Rupnik, Peter and Agnoloni, Tommaso and Aires, Jos{\'e} and Barkarson, Starkadur and Bartolini, Roberto and Bel, N{\'u}ria and Calzada P{\'e}rez, Mar{\'\i}a and Dar{\c{g}}is, Roberts and Diwersy, Sascha and Gavriilidou, Maria and van Heusden, Ruben and Iruskieta, Mikel and Kahusk, Neeme and Kryvenko, Anna and Ligeti-Nagy, No{\'e}mi and Magari{\~n}os, Carmen and M{\"o}lder, Martin and Navarretta, Costanza and Simov, Kiril and Tungland, Lars Magne and Tuominen, Jouni and Vidler, John and Vladu, Adina Ioana and Wissik, Tanja and Yrj{\"a}n{\"a}inen, V{\"a}in{\"o} and Fi{\v{s}}er, Darja},
    title = {ParlaMint II: Advancing Comparable Parliamentary Corpora across Europe},
    journal = {Language Resources and Evaluation},
    year = {2025},
    volume = {59},
    number = {3},
    pages = {2071--2102},
    issn = {1574-0218},
    doi = {10.1007/s10579-024-09798-w},
    url = {https://doi.org/10.1007/s10579-024-09798-w}
}

@inproceedings{10.1145/290941.291025,
    author = {Carbonell, Jaime and Goldstein, Jade},
    title = {The use of MMR, diversity-based reranking for reordering documents and producing summaries},
    year = {1998},
    isbn = {1581130155},
    publisher = {Association for Computing Machinery},
    address = {New York, NY, USA},
    url = {https://doi.org/10.1145/290941.291025},
    doi = {10.1145/290941.291025},
    booktitle = {Proceedings of the 21st Annual International ACM SIGIR Conference on Research and Development in Information Retrieval},
    pages = {335–336},
    numpages = {2},
    location = {Melbourne, Australia},
    series = {SIGIR '98}
}

@misc{walker1997,
      title={PARADISE: A Framework for Evaluating Spoken Dialogue Agents}, 
      author={Marilyn A. Walker and Diane J. Litman and Candace A. Kamm and Alicia Abella},
      year={1997},
      eprint={cmp-lg/9704004},
      archivePrefix={arXiv},
      primaryClass={cmp-lg},
      url={https://arxiv.org/abs/cmp-lg/9704004}, 
}

@article{Ji_2023,
    title={Survey of Hallucination in Natural Language Generation},
    volume={55},
    ISSN={1557-7341},
    url={http://dx.doi.org/10.1145/3571730},
    DOI={10.1145/3571730},
    number={12},
    journal={ACM Computing Surveys},
    publisher={Association for Computing Machinery (ACM)},
    author={Ji, Ziwei and Lee, Nayeon and Frieske, Rita and Yu, Tiezheng and Su, Dan and Xu, Yan and Ishii, Etsuko and Bang, Ye Jin and Madotto, Andrea and Fung, Pascale},
    year={2023},
    month=mar, pages={1–38} }

@techreport{Page1998PageRank,
    author = {Page, Lawrence and Brin, Sergey and Motwani, Rajeev and Winograd, Terry},
    institution = {Stanford Digital Library Technologies Project},
    title = {{The PageRank Citation Ranking: Bringing Order to the Web}},
    url = {http://citeseerx.ist.psu.edu/viewdoc/summary?doi=10.1.1.31.1768},
    year = 1998
}

@misc{edge2025localglobalgraphrag,
    title={From Local to Global: A Graph RAG Approach to Query-Focused Summarization}, 
    author={Darren Edge and Ha Trinh and Newman Cheng and Joshua Bradley and Alex Chao and Apurva Mody and Steven Truitt and Dasha Metropolitansky and Robert Osazuwa Ness and Jonathan Larson},
    year={2025},
    eprint={2404.16130},
    archivePrefix={arXiv},
    primaryClass={cs.CL},
    url={https://arxiv.org/abs/2404.16130}, 
}

@misc{liu2023evaluatingverifiabilitygenerativesearch,
    title={Evaluating Verifiability in Generative Search Engines}, 
    author={Nelson F. Liu and Tianyi Zhang and Percy Liang},
    year={2023},
    eprint={2304.09848},
    archivePrefix={arXiv},
    primaryClass={cs.CL},
    url={https://arxiv.org/abs/2304.09848}, 
}

@article{balogexpertise,
    author = {Balog, Krisztian and Fang, Yi and Rijke, Maarten and Serdyukov, Pavel and Si, Luo},
    year = {2012},
    month = {01},
    pages = {127-256},
    title = {Expertise Retrieval},
    volume = {6},
    journal = {Foundations and Trends in Information Retrieval},
    doi = {10.1561/1500000024}
}

@inproceedings{Zehlike_2017,
    series={CIKM ’17},
    title={FA*IR: A Fair Top-k Ranking Algorithm},
    url={http://dx.doi.org/10.1145/3132847.3132938},
    DOI={10.1145/3132847.3132938},
    booktitle={Proceedings of the 2017 ACM on Conference on Information and Knowledge Management},
    publisher={ACM},
    author={Zehlike, Meike and Bonchi, Francesco and Castillo, Carlos and Hajian, Sara and Megahed, Mohamed and Baeza-Yates, Ricardo},
    year={2017},
    month=nov, pages={1569–1578},
    collection={CIKM ’17}
}

@article{rashkin-etal-2023-measuring,
    title = "Measuring Attribution in Natural Language Generation Models",
    author = "Rashkin, Hannah  and
    Nikolaev, Vitaly  and
    Lamm, Matthew  and
    Aroyo, Lora  and
    Collins, Michael  and
    Das, Dipanjan  and
    Petrov, Slav  and
    Tomar, Gaurav Singh  and
    Turc, Iulia  and
    Reitter, David",
    journal = "Computational Linguistics",
    volume = "49",
    number = "4",
    month = dec,
    year = "2023",
    address = "Cambridge, MA",
    publisher = "MIT Press",
    url = "https://aclanthology.org/2023.cl-4.2/",
    doi = "10.1162/coli_a_00486",
    pages = "777--840"
}

@inproceedings{gao-etal-2023-rarr,
    title = "{RARR}: Researching and Revising What Language Models Say, Using Language Models",
    author = "Gao, Luyu  and
    Dai, Zhuyun  and
    Pasupat, Panupong  and
    Chen, Anthony  and
    Chaganty, Arun Tejasvi  and
    Fan, Yicheng  and
    Zhao, Vincent  and
    Lao, Ni  and
    Lee, Hongrae  and
    Juan, Da-Cheng  and
    Guu, Kelvin",
    editor = "Rogers, Anna  and
    Boyd-Graber, Jordan  and
    Okazaki, Naoaki",
    booktitle = "Proceedings of the 61st Annual Meeting of the Association for Computational Linguistics (Volume 1: Long Papers)",
    month = jul,
    year = "2023",
    address = "Toronto, Canada",
    publisher = "Association for Computational Linguistics",
    url = "https://aclanthology.org/2023.acl-long.910/",
    doi = "10.18653/v1/2023.acl-long.910",
    pages = "16477--16508"
}

@misc{hwang2024retrieval,
    title={Retrieval-Augmented Generation with Estimation of Source Reliability}, 
    author={Hwang, Jeongyeon and Park, Junyoung and Park, Hyejin and Kim, Dongwoo and Park, Sangdon and Ok, Jungseul},
    year={2024},
    eprint={2410.22954},
    archivePrefix={arXiv},
    primaryClass={cs.CL}
}

@article{hyvonen2025publishing,
  title={Publishing and using parliamentary Linked Data on the Semantic Web: ParliamentSampo system for Parliament of Finland},
  author={Hyv{\"o}nen, Eero and Sinikallio, Laura and Leskinen, Petri and Drobac, Senka and Leal, Rafael and La Mela, Matti and Tuominen, Jouni and Poikkim{\"a}ki, Henna and Rantala, Heikki},
  journal={Semantic Web},
  volume={16},
  number={1},
  pages={SW--243683},
  year={2025},
  publisher={SAGE Publications Sage UK: London, England}
}

@article{salamanca_kg_2024,
author = {Salamanca, Luis and Brandenberger, Laurence and Gasser, Lilian and Schlosser, Sophia and Balode, Marta and Jung, Vincent and Perez-Cruz, Fernando and Schweitzer, Frank},
title = {Processing Large-Scale Archival Records: The Case of the Swiss Parliamentary Records},
journal = {Swiss Political Science Review},
volume = {30},
number = {2},
pages = {140-153},
doi = {https://doi.org/10.1111/spsr.12590},
url = {https://onlinelibrary.wiley.com/doi/abs/10.1111/spsr.12590},
eprint = {https://onlinelibrary.wiley.com/doi/pdf/10.1111/spsr.12590},
year = {2024}
}

@misc{brandenberger2024democrasci,
  title={DemocraSci - A Parliamentary Knowledge Graph (4 legislative periods) (0.9.0) [Dataset]},
  author={Brandenberger, Laurence and Minder, Julian and Salamanca, Luis and Schlosser, Sophia and Gasser, Lilian and Jung, Vincent and Shariat, Kourosh and Balode, Marta and Schmidt-Rohr, Anna and Babi{\'c}, Leon and Perez-Cruz, Fernando and Schweitzer, Frank},
  year={2024},
  publisher={Zenodo},
  doi={10.5281/zenodo.13920293},
  url={https://doi.org/10.5281/zenodo.13920293}
}

@article{vanaggelen2017linkedep,
author = {van Aggelen, Astrid and Hollink, Laura and Kemman, Max and Kleppe, Martijn and Beunders, Henri},
editor = {Natasha Noy},
title = {The debates of the European Parliament as Linked Open Data},
year = {2017},
issue_date = {2017},
publisher = {IOS Press},
address = {NLD},
volume = {8},
number = {2},
issn = {1570-0844},
url = {https://doi.org/10.3233/SW-160227},
doi = {10.3233/SW-160227},
journal = {Semant. Web},
month = jan,
pages = {271–281},
numpages = {11}
}

@misc{daticamera2026,
  title={Portale dei dati aperti della Camera dei deputati},
  author={{Camera dei deputati}},
  howpublished={\url{https://dati.camera.it}},
  note={Accessed: 2026}
}

@InProceedings{bojars2019linkedsaeima,
author="Boj{\={a}}rs, Uldis
and Dar{\c{g}}is, Roberts
and Lavrinovi{\v{c}}s, Uldis
and Paikens, P{\={e}}teris",
editor="Acosta, Maribel
and Cudr{\'e}-Mauroux, Philippe
and Maleshkova, Maria
and Pellegrini, Tassilo
and Sack, Harald
and Sure-Vetter, York",
title="LinkedSaeima: A Linked Open Dataset of Latvia's Parliamentary Debates",
booktitle="Semantic Systems. The Power of AI and Knowledge Graphs",
year="2019",
publisher="Springer International Publishing",
address="Cham",
pages="50--56",
isbn="978-3-030-33220-4"
}

@misc{Mosbach2025,
  title={Parliamentary debates in The World Avatar: A hybrid retrieval-augmented generation system},
  author={Mosbach, Sebastian and Lai, Jiawei and Rustagi, Kushagar and Tran, Dan N and Kraft, Markus and Bindereif, Ewald and Azzam, Mark},
  year={2025},
  note={Preprint},
  url={https://como.ceb.cam.ac.uk/media/preprints/c4e-preprint-338.pdf}
}

@inproceedings{Azzopardi2025,
  author={Joel Azzopardi},
  title={LLMs and Knowledge Discovery in Low-Resource Language
Parliamentary Corpora: The PQ Dashboard Case Study},
  booktitle={Proceedings of the 17th International Joint Conference on Knowledge Discovery, Knowledge Engineering and Knowledge Management (IC3K 2025) - Volume 1: KDIR},
  pages={159--170},
  year={2025},
  publisher={SciTePress},
  doi={10.5220/0013835100004000},
  url={https://www.scitepress.org/Papers/2025/138351/138351.pdf}
}

\end{document}